\documentclass{llncs}

\usepackage{graphicx,wrapfig,lipsum,amsmath}
\usepackage{subcaption}	
\usepackage{multirow}
\usepackage{color,soul}
\usepackage[nocompress]{cite}

\begin{document}

\title{Shallow vs deep learning architectures for white matter lesion segmentation in the early stages of multiple sclerosis}

\author{******* \\ ****** }

\author{Francesco La Rosa\inst{1,3}\and M\'{a}rio Jo\~{a}o Fartaria \inst{1,2,4}
Tobias Kober\inst{1,2,4}\and Jonas Richiardi \inst{2,4} \and 
Cristina Granziera\inst{5,6} \and Jean-Philippe Thiran\inst{1,4} \and Meritxell Bach Cuadra\inst{1,3,4}}


\institute{********\\******}

\institute{LTS5, Ecole Polytechnique F\'{e}d\'{e}rale de Lausanne, Switzerland \and  Siemens Healthcare AG, Lausanne, Switzerland \and  Medical Image Analysis Laboratory, CIBM, University of Lausanne, Switzerland \and Radiology Department, Lausanne University Hospital, Switzerland \and Translational Imaging in Neurology Basel, Department of Medicine and Biomedical Engineering, University Hospital Basel and University of Basel, Basel,
Switzerland \and Neurologic Clinic and Policlinic, Departments of Medicine, Clinical Research and Biomedical Engineering, University Hospital Basel and University of Basel, Basel,
Switzerland}

\maketitle              

\begin{abstract}
In this work, we present a comparison of a shallow and a deep learning architecture for the automated segmentation of white matter lesions in MR images of multiple sclerosis patients. In particular, we train and test both methods on early stage disease patients, to verify their performance in challenging conditions, more similar to a clinical setting than what is typically provided in multiple sclerosis segmentation challenges. Furthermore, we evaluate a prototype naive combination of the two methods, which refines the final segmentation. All methods were trained on 32 patients, and the evaluation was performed on a pure test set of 73 cases. Results show low lesion-wise false positives (30\%) for the deep learning architecture, whereas the shallow architecture yields the best Dice coefficient (63\%) and volume difference (19\%). Combining both shallow and deep architectures further improves the lesion-wise metrics (69\% and 26\% lesion-wise true and false positive rate, respectively).
\end{abstract}

\section{Introduction}
\label{sec:intro}
Multiple Sclerosis (MS) is a demyelinating disease that affects the central nervous system. Demyelination results in focal lesions that appear with higher frequency in the white matter (WM). Magnetic resonance imaging (MRI) is a fundamental tool for MS diagnosis and monitoring of disease evolution as well as response to therapy. Currently, expert's manual annotations are considered the clinical gold standard for MS lesion identification. However, as this task is time-consuming and prone to inter and intra-observer variations, many automated methods for MS lesion detection and segmentation have been proposed in the literature~\cite{review1}.
In this context, supervised techniques that learn and train from manually annotated examples have been proven to be the most successful in detection of MS WM lesions ~\cite{challenge2008,challenge2015,miccai2016}. In the last years, deep learning architectures have achieved remarkable successes and have recently proven good performance in MS lesion segmentation as well~\cite{Brosch,Valverde,Roy}.

In order to compare automated lesion segmentation methods, several computational imaging challenges have been proposed at international conferences~\cite{challenge2008,challenge2015,miccai2016}, providing very valuable benchmark datasets for validation. However, these evaluation scenarios are based on patients with relatively high lesion load, and reported results are often computed on scans exhibiting relative large lesion sizes. Thus, the performance of deep supervised techniques on early stages of MS and at small lesion sizes remains to be proven.

In this work, we aim at comparing shallow with novel deep learning architectures using data from early stages of the disease in challenging conditions, i.e. exploring minimum lesion sizes as given by neuroradiological conventions~\cite{Grahl} and even pushing the limit below.

To this end, we have selected two recently published MS segmentation methods. First, we have applied a supervised k-NN method combined with partial volume (PV) modeling \cite{Fartaria,FartariaMiccai}, specifically developed on subjects with a low disease burden and small lesions. Second, we have used a recently and publicly available deep learning approach based on a cascade of two 3D patch-wise convolutional neural networks (CNNs) \cite{Valverde}. At the time of the writing of this work, this CNNs method achieved the best result on the MICCAI2008 and MSSEG2016 challenges \cite{challenge2008,miccai2016} and competitive performance on other clinical datasets. Furthermore, we explore a straightforward prototype combination of these two methods. Both methods and their combination are trained on the same clinical dataset and validated on a pure test set. The results are analyzed considering different minimum lesion volume and total lesion load, as these are important evidences for early stage disease patients with low disabilities.

\section{Methodology} 
\label{sec:method}

\subsection{Datasets}
\label{datasetSection}
The study was approved by the Ethics Committee of our institution, and all patients gave written informed consent prior to participation. The training dataset was composed of 32 patients, 18 female / 14 male, mean age $34 \pm 10$ years, with Expanded Disability Status Scale (EDSS) scores ranged from 1 to 2 (mean $1.6 \pm 0.3$). Mean lesion volume is $0.11 \pm 0.40$ ml (range 0.001-7.03 ml). Mean lesion load per case was $6.0 \pm 7.2$ ml (range 0.3-37.2 ml). MRI acquisitions were performed on a 3T MRI scanner (Magnetom Trio, Siemens Healthcare, Erlangen, Germany). Both 3D MPRAGE and 3D FLAIR were acquired with a resolution of 1 x 1 x 1.2 mm$^{3}$.

The test dataset was made up of 73 patients, 50 females and 23 males (mean age $38 \pm 10$ years). EDSS scores ranged from 1 to 7.5 (mean $2.6 \pm 1.5$). Mean lesion volume was $0.25 \pm 3.29$ ml (range 0.002-159.827 ml). Mean lesion load per case was $14.3 \pm 27.9$ ml (range 0.2-162.9 ml). Both 3D MPRAGE and 3D FLAIR were acquired at 1 x 1 x 1 mm$^{3}$ but with different Siemens scanners: 5 subjects at 1.5T with MAGNETOM Aera, and the other patients at 3T with either Prisma\_fit, TrioTim, or Skyra systems.

\textbf{Manual segmentation:}
In the training set, MS lesions were detected by consensus by one radiologist and one neurologist, with respectively 6 and 11 years of experience. The lesion volumes were then delineated in each image by a trained technician.
Testing set lesions segmentation was performed by the Medical Image Analysis Center-MIAC \cite{miac} based on a standardized semi-automated method and further experts quality check, which has been extensively applied to phase II and III clinical trials.

\subsection{Pre-processing}
The same pre-processing steps were applied to the training and testing datasets. First, the two image contrasts were rigidly registered to the same space (MPRAGE) using the ELASTIX C++ library \cite{elastix}. Second, all cases were skull-stripped using BET \cite{bet} and bias-corrected using N4~\cite{n4,3dslicer}.

\subsection{LeMan-PV} 
LeMan-PV is a Bayesian PV estimation (PVE) algorithm, where spatial constraints for GM and lesions are included to drive the segmentation~\cite{FartariaMiccai}. The spatial constraint for GM is an atlas-based probability map, and spatial constraints for lesions are derived from a kNN-supervised-based approach \cite{Fartaria3}. LeMan-PV has proven its good performance, and improvements as compared to state-of-the-art methods, in a leave-one-out experiments with MS patients with low lesion loads and small lesions. As in~\cite{Fartaria}, initial mean tissue intensities and hyperparameters (symmetric penalty matrix $A$, and amount  of  spatial  smoothness $\beta$) were set and a patient with relatively high lesion load chosen as a reference to train the PV estimation algorithm. Specifically, $A$ coefficients were $a_1 = 11.25$, $a_4 = 14.33$, $a_5 = 0.47$, $a_6 = 12.21$, $a_7 = 1.33$, $a_8 = 16.93$, and $\beta = 0.5$. Patient mean intensities were set determined beforehand by histogram matching with the same reference patient used for hyperparameter setting~\cite{intensity}.

\subsection{CNNs}
A novel MS segmentation method based on a cascade of two 3D patch-wise CNNs has recently been proposed \cite{Valverde}. The two networks have the same architecture and number of parameters, but don't share the same weights. Added to the above pre-processing steps, additional intensity normalization was performed, applying a histogram matching technique \cite{intensity}. Afterwards, the first CNN receives as input patches of size 11x11x11 from different MRI modalities, centered around a voxel of interest. Only voxels with a FLAIR intensity over a threshold optimized in the validation phase are considered. Lesion candidates from the first CNN are then given as input to the second one, which mainly has the task of reducing the false positives. In order to overcome the problem of data imbalance, before each CNN the negative class is undersampled, and the same number of positive and negative patches are obtained. Binary output masks are computed by linearly thresholding the probabilistic lesion masks given as output by the second network.
\setlength{\tabcolsep}{10pt}
\begin{table}[ht!]
\small
\centering
\caption{Network architecture. c indicates the number of MRI modalities.}\label{tab1}
\begin{tabular}{|l l l l|}
\hline
Layer &  Type  & Output size &  Feature maps\\
\hline
0 & Input & c x 11 x 11 x 11 & -\\
1 & Convolutional & 32 x 11 x 11 x 11 & 32\\
2 & Convolutional & 32 x 11 x 11 x 11 & 32\\
3 & Max-pooling  & 32 x 5 x 5 x 5 & -\\
4 & Convolutional & 64 x 5 x 5 x 5 & 64\\
5 & Convolutional & 64 x 5 x 5 x 5 & 64\\
6 & Max-pooling & 64 x 2 x 2 x 2 & -\\
7 & FC & 256 & 256\\
8 & Softmax & 2 & 2\\
\hline
\end{tabular}
\label{tab1:architecture}
\end{table}

We have applied the same architecture~\cite{Valverde} publicly available at~\cite{valgithub} (see Table \ref{tab1:architecture}). Each convolutional layer is followed by a ReLU activation function and a batch normalization regularization. Dropout (p=0.5) is applied before the first fully-connected layer. The networks were trained with the adaptive learning rate method (ADADELTA) \cite{Adadelta}, a batch size of 128, and early stopping as in the original paper. From the training dataset 7 cases were kept for validation, leaving 25 cases for training. With these the binarisation threshold was optimized considering equally the dice coefficient and the lesion false positive rate.
In the original work \cite{Valverde} the CNNs were trained with 20 to 35 cases. Therefore, having a comparable number of patients for training, we hypothesize that this method should not perform worse in our study.  
\subsection{Combination of LeMan-PV with CNNs}
\label{comb}
It has been shown that for segmentation tasks, CNNs can benefit from prior probability maps fed in as an additional input channel \cite{Luo,Zotti}. Moreover, combining different classifiers has also been a successful technique for improving the final results in supervised learning in several works \cite{combining,Fartaria4,Ensemble}. Here, we propose a naive prototype combination (PV-CNNs) of both approaches described above. The concentration lesion maps generated by LeMan-PV are included as an additional input channel of the first CNN during training and testing. In this way, additional prior information on lesions was given to the network with the aim of improving the final segmentation.

\section{Results}
\label{eval}
\begin{figure}[ht!]
	\includegraphics[width=.23\linewidth]{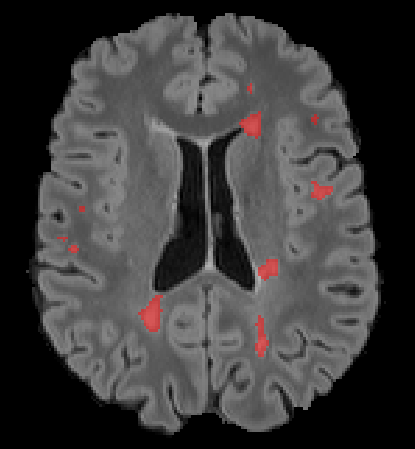}	
	\includegraphics[width=.23\linewidth]{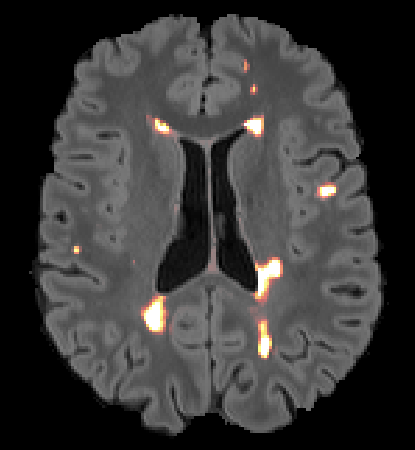}
	\includegraphics[width=.23\linewidth]{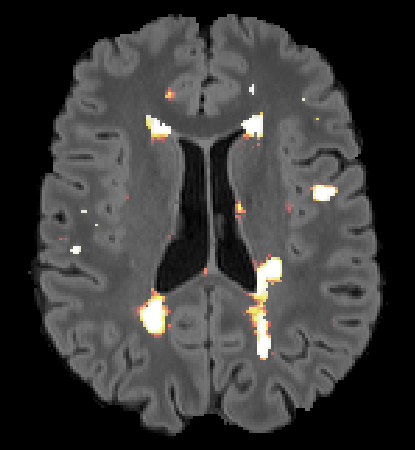}
	\includegraphics[width=.23\linewidth]{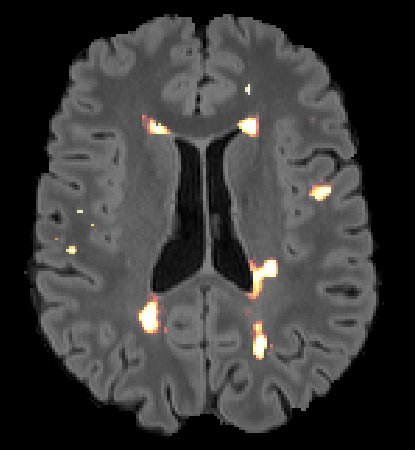}
    \includegraphics[height=.25\linewidth]{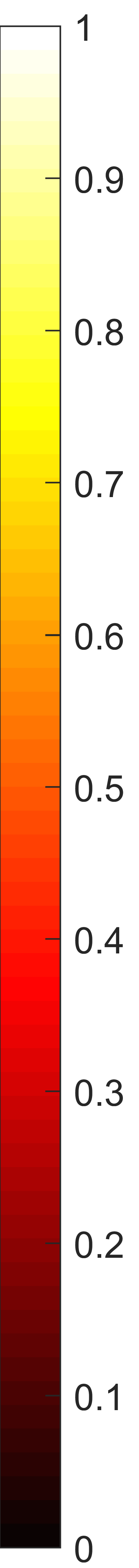}
\caption{Segmentation results (lesion probability (CNNs, PV-CNNs) and concentration (LeMan-PV)), from left to right: ground truth, LeMan-PV, CNNs, PV-CNNs. Reduction of LFPR is observed in PV-CNNs.\label{examples4}}
\end{figure}

We compared the results of LeMan-PV, CNNs, and PV-CNNs strategies (see Figure \ref{examples4}). In line with three MS lesion segmentation challenges~\cite{challenge2008,challenge2015,miccai2016}, we computed the following evaluation metrics: overlap Dice coefficient (Dice), lesion-wise false positive (LFPR) and lesion-wise true positive (LTPR) rates, voxel-wise true positives (TP), and volume difference (VD), according to~\cite{challenge2015,metrics}. Rather than a leave-one-out analysis~\cite{Valverde,FartariaMiccai}, we present our results on a pure testing set of 73 patients cases acquired with different scanners. These two factors allow us to evaluate the generalization of the proposed methods in a setting close to the clinical scenario (shown in Table~\ref{tab3}).

\begin{figure}[ht!] 
	\includegraphics[width=.334\linewidth]{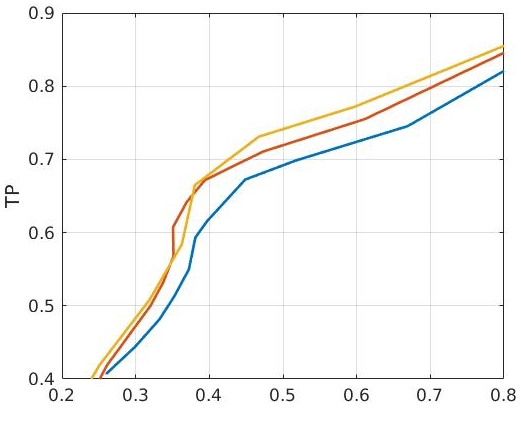}	
	\includegraphics[width=.320\linewidth]{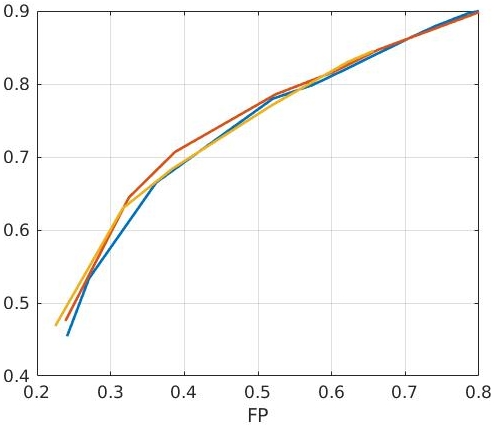}
	\includegraphics[width=.330\linewidth]{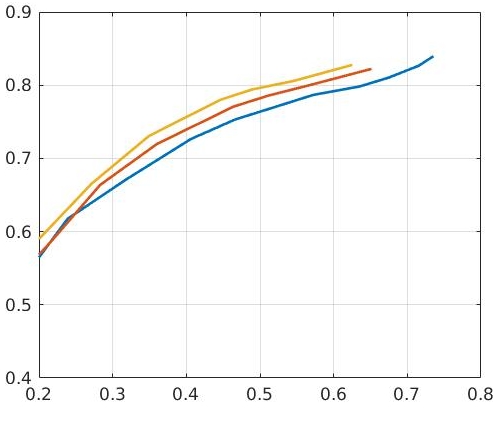}
\caption{ROC curves for different minimum lesion size: 5 (blue), 10 (orange), 15 (yellow) mm$^{3}$. From left to right: LeMan-PV, CNNs, PV-CNNs.}
\label{roc}
\end{figure}

Quantitative evaluation at different lesion sizes (5, 10, 15 mm$^{3}$) is given by ROC curves in Figure \ref{roc}. Both LeManPV and PV-CNNs performed better at bigger minimum lesion size. However, CNNs did not show this behavior in our cohort, presenting similar ROC curves for all minimum lesion sizes. 

As in the original studies~\cite{Fartaria,Valverde}, in what follows, a minimum lesion size of 5 mm$^{3}$ is considered. Median values for the whole test dataset are reported in Table~\ref{tab2}. LeManPV achieved the best Dice coefficient and volume difference. However, in terms of LFPR and LTPR, the CNNs performed better. The combination of the two methods outperformed them singularly in these lesion-wise metrics.

\setlength{\tabcolsep}{12pt}
\begin{table}
\centering
\caption{Median values of the evaluation metrics for each method considered.}\label{tab2}
\begin{tabular}{|l | l l l l l|}
\hline
Method &  Dice  & LFPR &  LTPR & TP & VD\\
\hline
LeMan-PV & \textbf{0.63} & 0.37 & 0.57 & \textbf{0.66} & \textbf{0.19}\\
CNNs & 0.57 & 0.30 & 0.66 & 0.56 & 0.26\\
PV-CNNs & 0.60 & \textbf{0.26} & \textbf{0.69} & \textbf{0.66} & 0.40\\
\hline
\end{tabular}
\label{tab2}
\end{table}

\begin{figure} [h!]
  \centering
  \includegraphics[width=6cm]{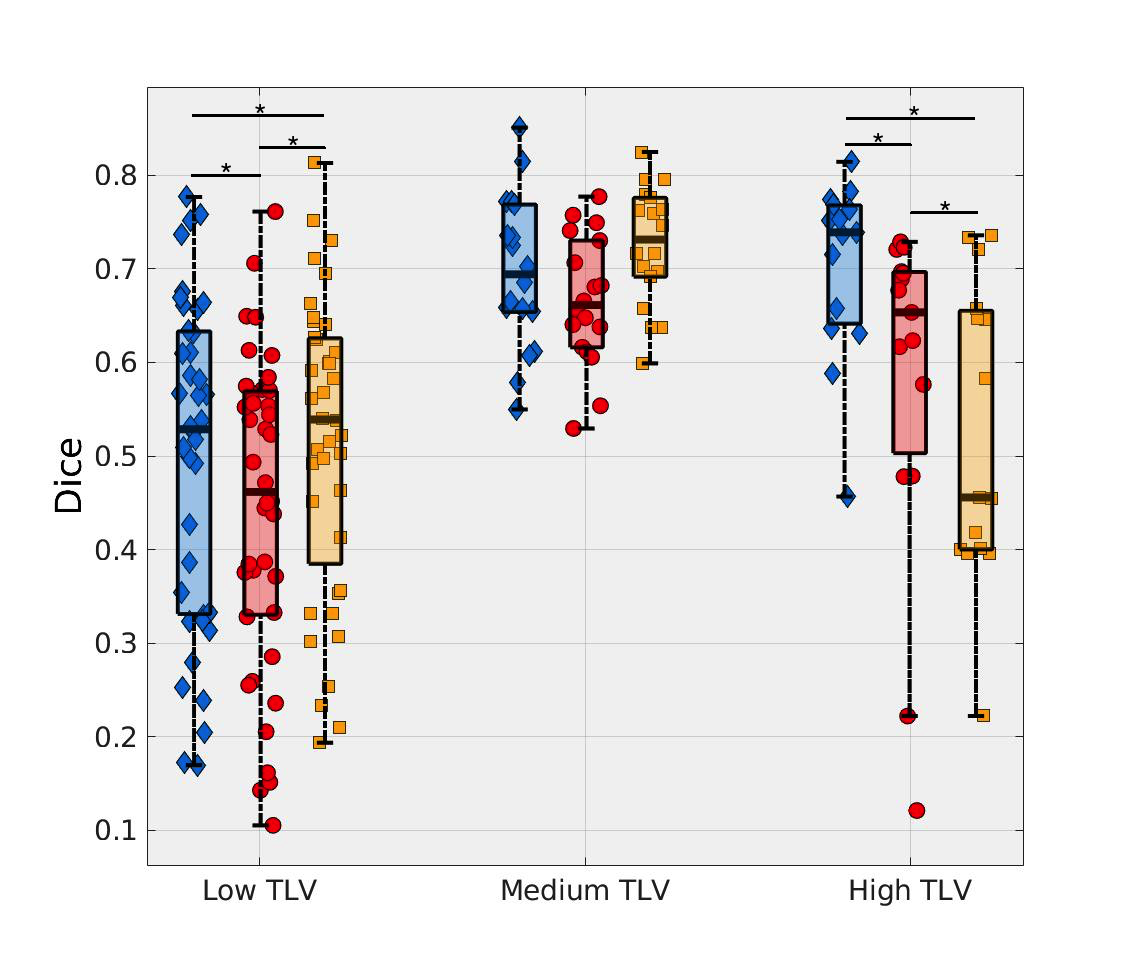}
  \includegraphics[width=6cm]{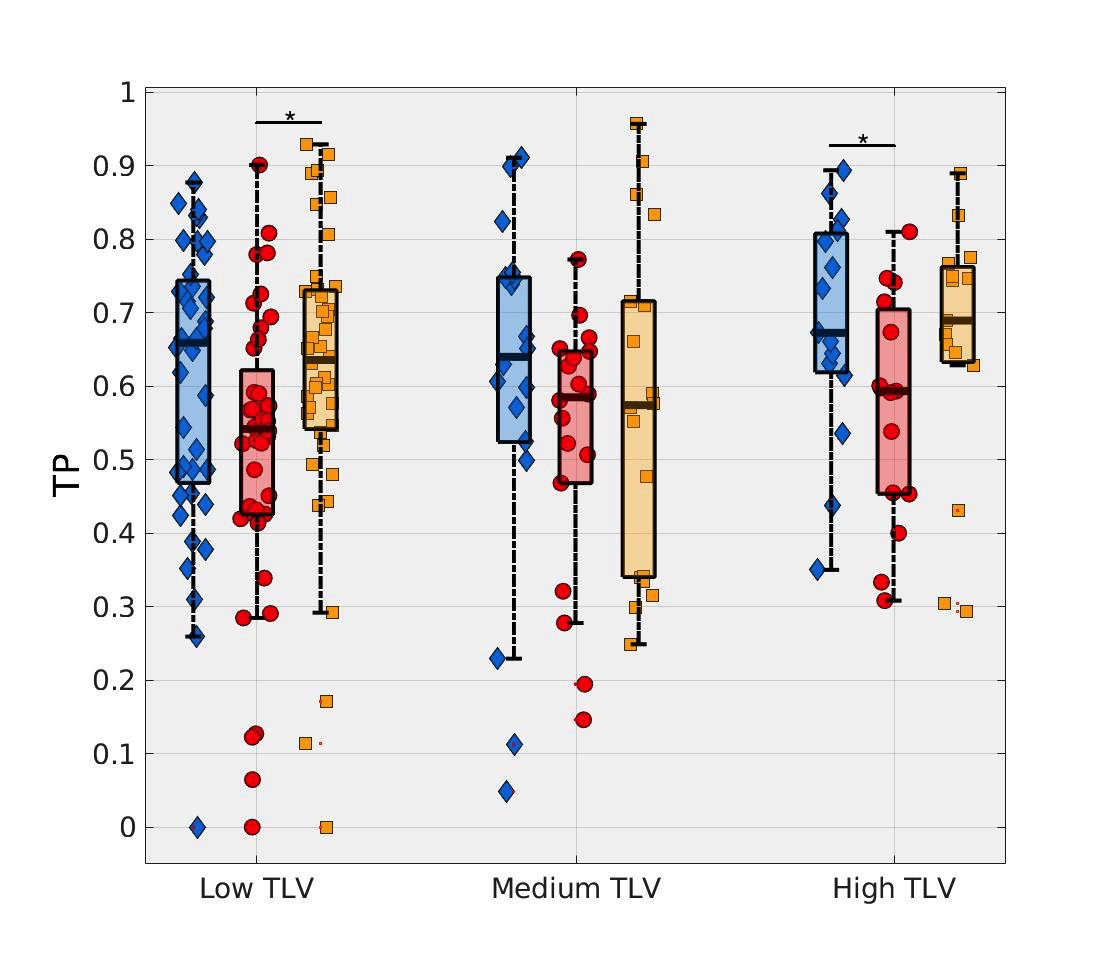}
  \includegraphics[width=6cm]{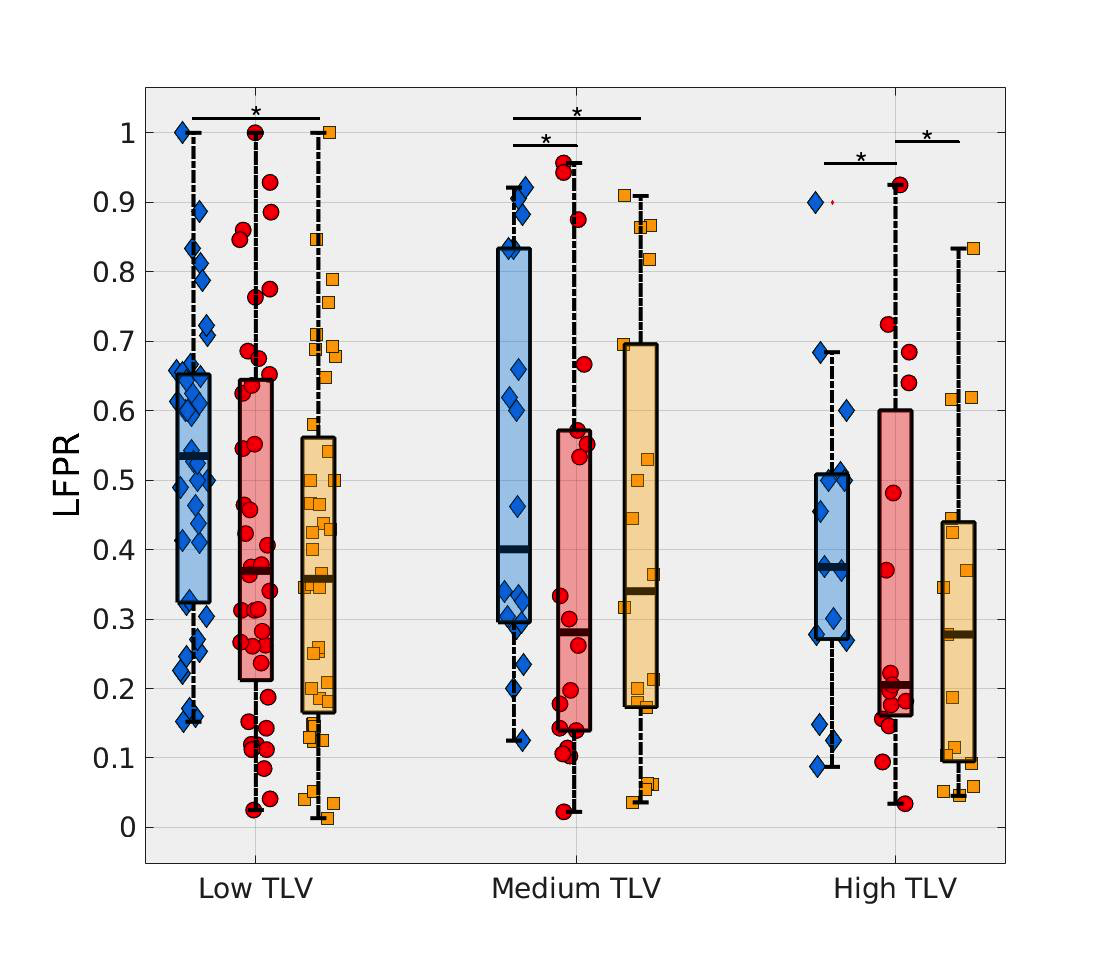}
  \includegraphics[width=6cm]{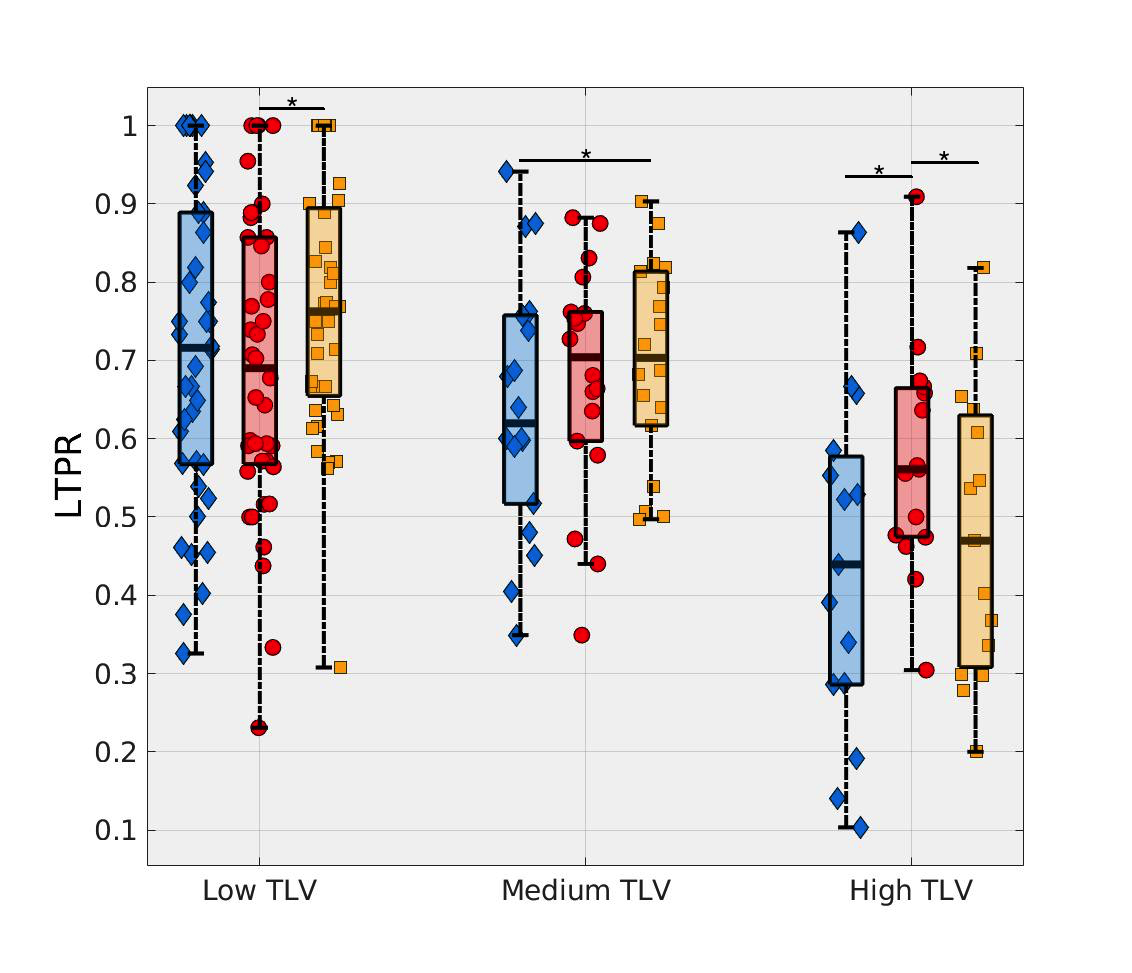}
  \includegraphics[width=1.5cm]{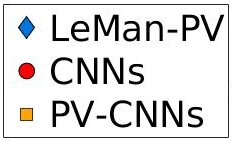}
  \caption{Box plots of the Dice coefficient, TP, LFPR, and LTPR for the three methods considering the different TLV of the testing cases ($p<0.05$ is indicated by *).}
  \label{fig1:Dice_box}
\end{figure}

Segmentation results by Dice coefficient, TP, LFPR, and LTPR are given in the boxplots of Figure~\ref{fig1:Dice_box}. Results are split in groups of patients according to their total lesion volume (TLV). In agreement with \cite{Fartaria3}, we considered a low (TLV $<$ 5 $ml$), moderate (5 $ml$ $\leq$ TLV $\leq$ 15 $ml$) and high (TLV $>$ 15 $ml$) total lesion burden. Statistically significant differences between the methods are computed with Wilcoxon signed-rank test (p \textless 0.05 uncorrected). Interestingly, PV-CNNs achieved the best Dice coefficient for low and medium TLV, but its performance drops for high lesion load. We hypothesize that the lower number of cases in this category (only 15 patients) downgrades the classification results for CNNs weakness to statistics. Overall, besides the presence of some outliers, LeMan-PV and CNNs showed a similar behavior at low and medium lesion loads. Regarding the TP, there are not significant differences between the three TLV. On the other hand, the LFPR decreases for all methods as the TLV increases. This represents an understandable behavior, as higher lesion load cases are expected to be better segmented. Curiously, and opposite to TP, the LTPR follows a similar trend. However, as stated above, the low number of patients at highest lesion load prevents us from drawing conclusions.

\begin{figure}[ht!]
	\includegraphics[width=.332\linewidth]{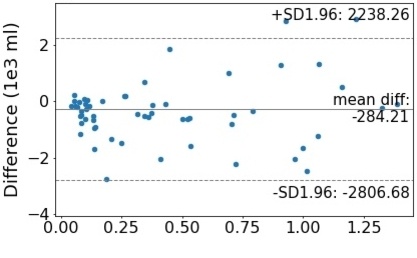}	
	\includegraphics[width=.318\linewidth]{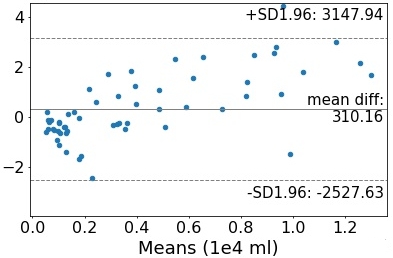}
	\includegraphics[width=.323\linewidth]{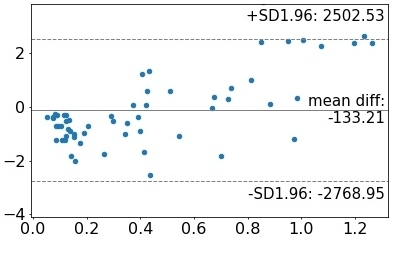}
    \includegraphics[width=.332\linewidth]{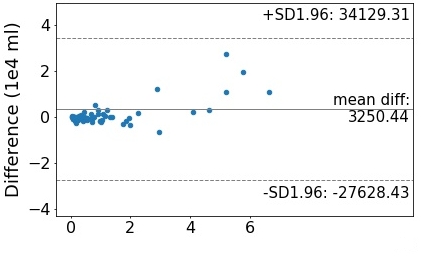}	
	\includegraphics[width=.318\linewidth]{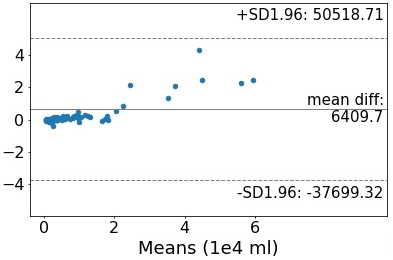}
	\includegraphics[width=.323\linewidth]{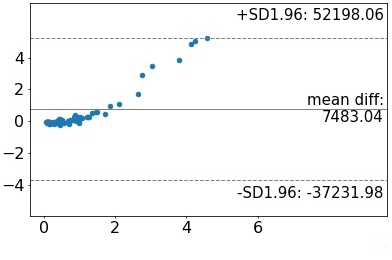}
\caption{Bland-Atlmann plots of low and medium TLV cases (top row) and of the whole dataset (bottom row) showing the volume differences of the three methods analyzed. From left to right: LeMan-PV, CNNs, PV-CNNs.}
\label{bland}
\end{figure}

Volume differences are given (top row, for low and medium TLV patients only, bottom row: all dataset) by Bland-Altman plots (Figure \ref{bland}). Slightly better results were obtained when combining both architectures, with a mean volume difference of -133.21 $ml$. However, a different behavior is shown when including the high TLV patients, with an increase of the mean volume difference to 3250, 6410 and 7483 $ml$ for LeMan-PV, CNNs and PV-CNNs respectively. 

Finally, the effect of the scanner type is briefly investigated. Table \ref{tab3} shows the mean Dice coefficient for the four different scanner types used to acquire the testing cases. For all segmentation methods, the highest Dice coefficient is achieved for the cases acquired with the TrioTim scanner. However, in this work the number of cases for each scanner is highly unbalanced. Therefore, further studies, with enlarged datasets, will be needed to quantify accuracy versus scanner type.  

\setlength{\tabcolsep}{9pt}
\begin{table}
\centering
\caption{Mean Dice coefficient of the testing cases for the different scanners they were acquired with.}
\begin{tabular}{c  c | c  c  c|}
\cline{3-5}
& & \multicolumn{3}{|c|}{Dice (range)} \\ 
\cline{1-5}
\multicolumn{1}{|c|}{Scanner} &  N. cases  & LeMan-PV & CNNs & PV-CNNs\\
\hline
\multicolumn{1}{|c|}{Aera} & \ 5 &  0.59 (0.47-0.64) & 0.47 (0.12-0.75) & 0.52 (0.29-0.63)\\
\multicolumn{1}{|c|}{TrioTim}  & 6 & 0.63 (0.50-0.81) & 0.61 (0.44-0.75) & 0.65(0.51-0.80)\\
\multicolumn{1}{|c|}{Prisma\_fit} & 11 & 0.54 (0.31-0.74) & 0.48 (0.26-0.64) & 0.53 (0.34-0.72)\\
\multicolumn{1}{|c|}{Skyra} & 51 & 0.59 (0.16-0.84) & 0.53 (0.11-0.78)  & 0.56 (0.19-0.80)\\
\hline
\end{tabular}
\label{tab3}
\end{table}

\section{Conclusion}
\label{sec4}
In this work, we presented the comparison of two of the most recent automated methods for WM lesions segmentation published in literature. In particular, we have tested a Bayesian partial volume estimation algorithm (LeMan-PV) \cite{Fartaria} and a novel deep learning architecture based on a cascade of CNNs \cite{Valverde}. Both methods were tested on a pure test dataset composed of 73 cases, mainly belonging to early stage disease patients. The CNNs achieved the lowest LFPR of 30\%. This confirms, as claimed in the original paper \cite{Valverde}, that they are an effective method for reducing false positives. However, LeMan-PV showed the best segmentation results with the highest Dice coefficient (63\%) and smallest volume difference (19\%), indicating that PV might be still an asset for good delineation. Further analysis indicates a slight dependence of LeMan-PV performance on the minimum lesion size considered, whereas the CNNs didn't show this behavior. Furthermore, a combination of the two methods (PV-CNNs) was implemented. Providing the CNNs with the probability maps of the LeMan-PV improved the LFPR (26\%) and LTPR (69\%) but did not perform well in terms of VD. Those results confirm that the hybrid of the two methods is also effective for WM lesion segmentation of early stages disease cases. However, further improvements are needed to increase the segmentation accuracy of low lesion burden cases, in which these automated methods achieved the worst performance (median Dice around 0.5). These cases are indeed of great importance for detecting MS lesions in the early stages of the disease. Future work will include experimenting with advanced combinations of these methods, training and testing on different datasets, and verifying if the results depend on the scanner used.  

\section*{Acknowledgements}
The work is supported by the Centre d$'$Imagerie BioM\'{e}dicale (CIBM) of the University of Lausanne (UNIL), the Swiss Federal Institute of Technology Lausanne (EPFL), the University of Geneva (UniGe), the Centre Hospitalier Universitaire Vaudois (CHUV), the H\^{o}pitaux Universitaires de Gen\`{e}ve (HUG), and the Leenaards and Jeantet Foundations. This project is also supported by the European Union's Horizon 2020 research and innovation program under the Marie Sklodowska-Curie project TRABIT (agreement No 765148). CG is supported by the Swiss National Science Foundation grant SNSF Professorship PP00P3-176984.

\end{document}